\documentclass[10pt,twocolumn,letterpaper]{article}

\usepackage{iccv}
\usepackage{times}
\usepackage{epsfig}
\usepackage{graphicx}
\usepackage{amsmath}
\usepackage{amssymb}
\usepackage{placeins}

\newcommand{\fig}[1]{Figure~\ref{fig:#1}}

\newcommand{\tab}[1]{Table~\ref{tab:#1}}

\newcommand{\eq}[1]{(\ref{eq:#1})}

\usepackage{float}
\usepackage{caption}

\usepackage[accsupp]{axessibility}  % Improves PDF readability for those with disabilities.

\usepackage[pagebackref=true,breaklinks=true,letterpaper=true,colorlinks,bookmarks=false]{hyperref}
\iccvfinalcopy

\newcommand\fnurl[2]{%
  \href{#2}{#1}\footnote{\url{#2}}%
}

\def\iccvPaperID{8840} % *** Enter the ICCV Paper ID here

\ificcvfinal\fi

\begin{document}

%%%%%%%%% TITLE
\title{Point-Based Modeling of Human Clothing}

\author{
Ilya Zakharkin $^{1,2}$\thanks{* denotes equal contribution. VL is currently with Yandex and Skoltech.}   , \;
% \and
Kirill Mazur $^{1}$\footnotemark[1]  , \;
% \and
Artur Grigorev $^1$, \;
% \and
Victor Lempitsky $^{1,2}$ \\
% \and
% \textit{lempitsky@skoltech.ru}
\\ $^1$ Samsung AI Center, Moscow 
\\ $^2$ Skolkovo Institute of Science and Technology (Skoltech), Moscow
% \thanks{VL is currently with Yandex and Skoltech.}
\vspace{-20pt}
}

\twocolumn[{%
\renewcommand\twocolumn[1][]{#1}%
\maketitle
\begin{center}
    \centering
        \setlength{\tabcolsep}{.05em}
        \centering
        \vspace{0.cm}
        \newcommand{\tsrpcd}[2]{
                \includegraphics[trim={#2 0 #2 0},clip,height=4.6cm]{teaser/#1_pcd_0.002.png} 
                
            }
        \newcommand{\tsrrgb}[2]{
                \includegraphics[trim={#2 0 #2 0},clip,height=4.6cm]{teaser/#1_rgb.png} 
                
            }
    
        \centering
        \begin{tabular}{cccccccc}
             \tsrpcd{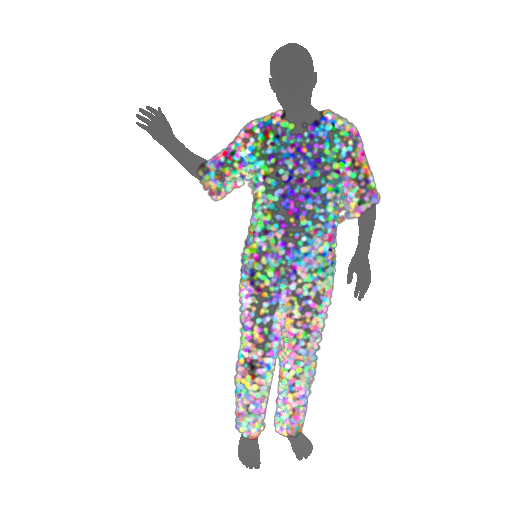}{130} &  \tsrpcd{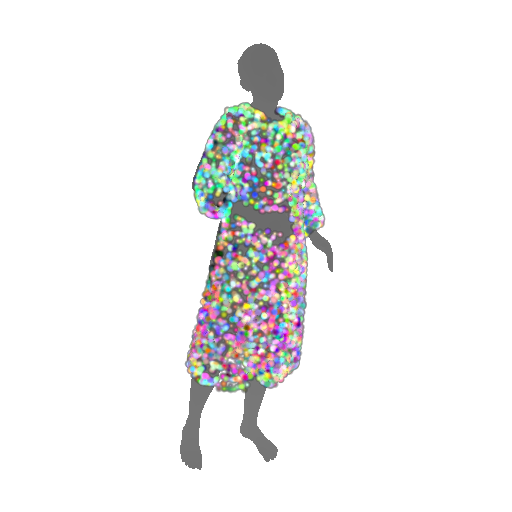}{150}  & 
             \tsrpcd{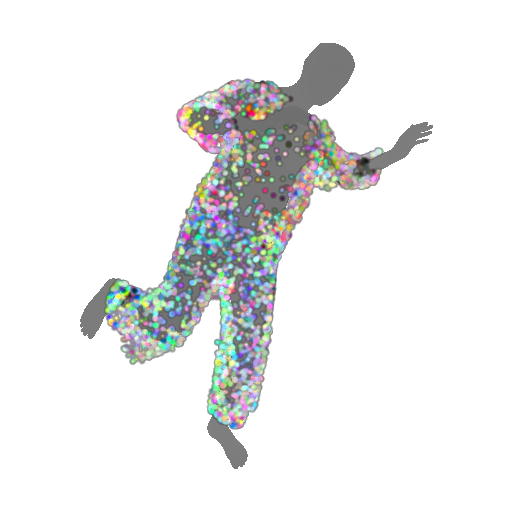}{90} &
             \tsrpcd{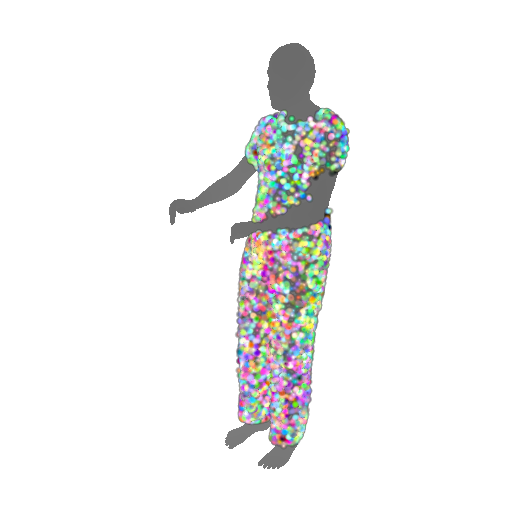}{165} &  \tsrpcd{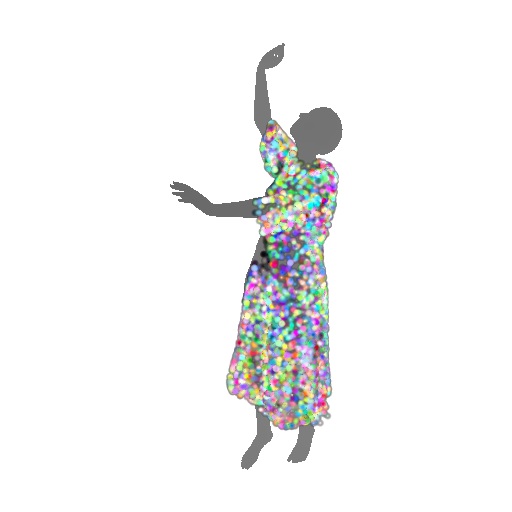}{150} &  \tsrpcd{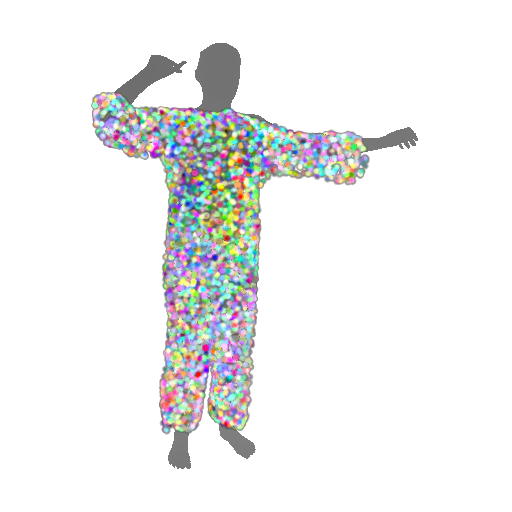}{100} &  \tsrpcd{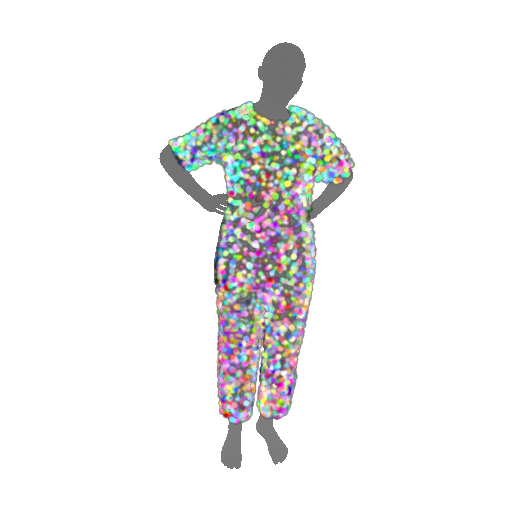}{150} &  \tsrpcd{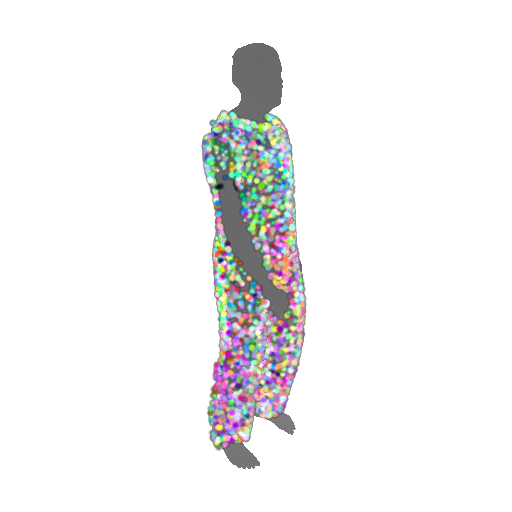}{150} \\
             \tsrrgb{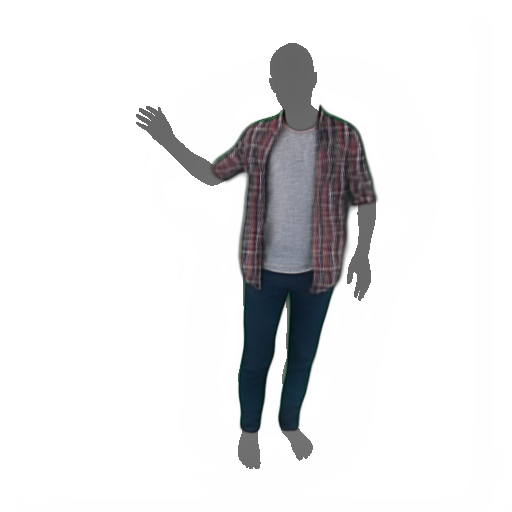}{130} &  \tsrrgb{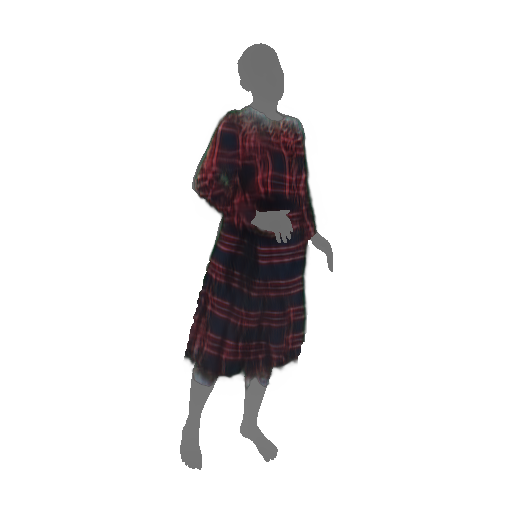}{150}  & 
             \tsrrgb{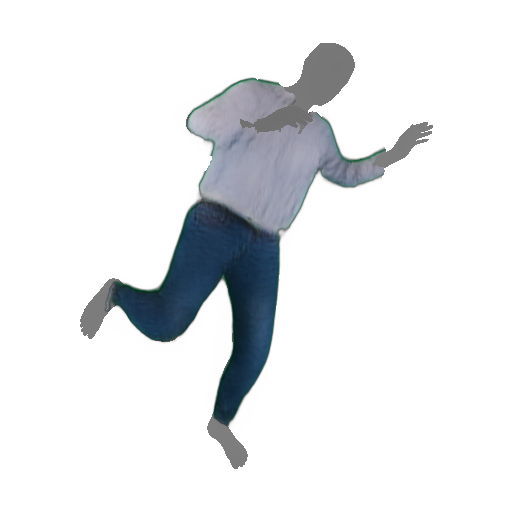}{90} &
             \tsrrgb{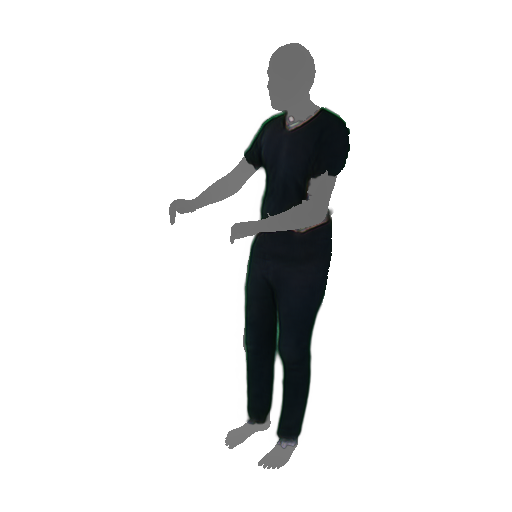}{165} &
             \includegraphics[trim={150 0 150 0},clip,height=4.6cm]{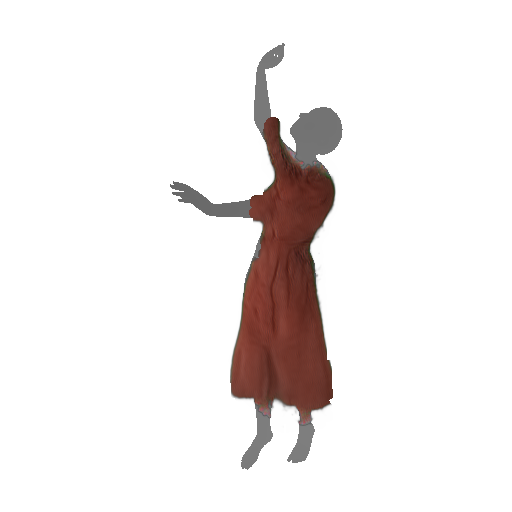} &
             \tsrrgb{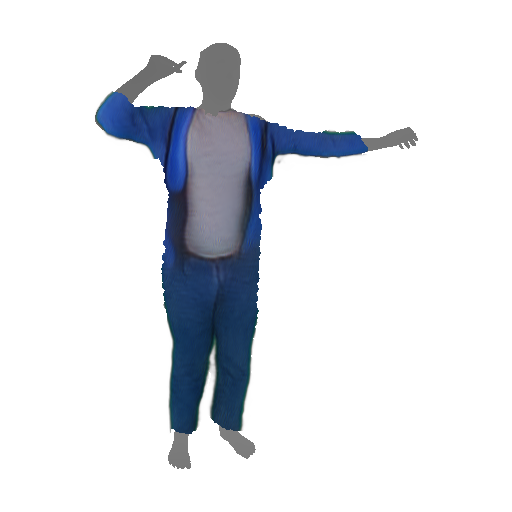}{100} &  \tsrrgb{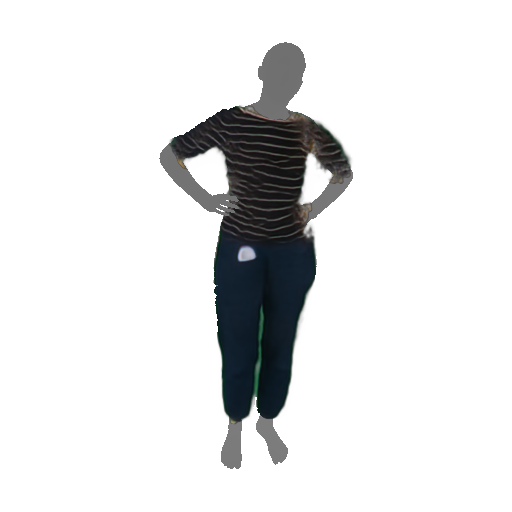}{150} &  \tsrrgb{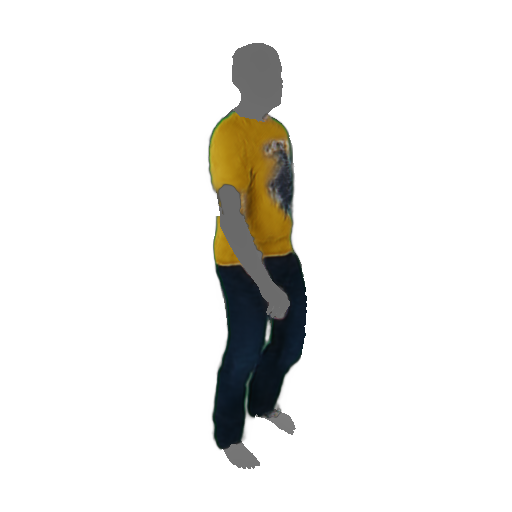}{150} \\ 
        \end{tabular}
        %\centering{
        \captionof{figure}{Our approach models the geometry of diverse clothing outfits using point clouds (top row; random point colors). The point clouds are obtained by passing the SMPL meshes (shown in grey) and latent outfit code vectors through a pretrained deep network. Additionally, our approach can model clothing appearance using \textit{neural point-based graphics} (bottom row). The outfit appearance can be captured from a video sequence, while a single frame is sufficient for point-based geometric modeling. }%}
        \label{fig:garment}
\end{center}%
}]

{
  \renewcommand{\thefootnote}%
    {\fnsymbol{footnote}}
  \footnotetext[1]{denotes equal contribution. VL is currently with Yandex and Skoltech.}
}

% \twocolumn[
% \begin{@twocolumnfalse}
\maketitle
\thispagestyle{empty}

% \end{@twocolumnfalse}
% ]
\begin{abstract}
We propose a new approach to human clothing modeling based on point clouds. Within this approach, we learn a deep model that can predict point clouds of various outfits, for various human poses, and for various human body shapes. Notably, outfits of various types and topologies can be handled by the same model. Using the learned model, we can infer the geometry of new outfits from as little as a single image, and perform outfit retargeting to new bodies in new poses. We complement our geometric model with appearance modeling that uses the point cloud geometry as a geometric scaffolding and employs neural point-based graphics to capture outfit appearance from videos and to re-render the captured outfits. We validate both geometric modeling and appearance modeling aspects of the proposed approach against recently proposed methods and establish the viability of point-based clothing modeling.
\end{abstract}

% \saythanks

% \maketitle
% \thispagestyle{empty}
% \begin{abstract}
% We propose a new approach to human clothing modeling based on point clouds. Within this approach, we learn a deep model that can predict point clouds of various outfits, for various human poses, and for various human body shapes. Notably, outfits of various types and topologies can be handled by the same model. Using the learned model, we can infer the geometry of new outfits from as little as a single image, and perform outfit retargeting to new bodies in new poses. We complement our geometric model with appearance modeling that uses the point cloud geometry as a geometric scaffolding and employs neural point-based graphics to capture outfit appearance from videos and to re-render the captured outfits. We validate both geometric modeling and appearance modeling aspects of the proposed approach against recently proposed methods and establish the viability of point-based clothing modeling.
% \end{abstract}

\section{Introduction}
Modeling realistic clothing is a big part of the overarching task of realistic modeling of humans in 3D. Its immediate practical applications include virtual clothing try-on as well as enhancing the realism of human avatars for telepresence systems. Modeling clothing is difficult since outfits have wide variations in geometry (including topological changes) and in appearance (including wide variability of textile patterns, prints, as well as complex cloth reflectance). Modeling interaction between clothing outfits and human bodies is an especially daunting task.

In this work, we propose a new approach to modeling clothing (\fig{garment}) based on point clouds. Using a recently introduced synthetic dataset~\cite{bertiche2020cloth3d} of simulated clothing, we learn a joint geometric model of diverse human clothing outfits. The model describes a particular outfit with a latent code vector (the \textit{outfit code}). For a given outfit code and a given human body geometry (for which we use the most popular SMPL format~\cite{smpl}),  a deep neural network (the \textit{draping network}) then predicts the point cloud that approximates the outfit geometry draped over the body. 

The key advantage of our model is its ability to reproduce diverse outfits with varying topology using a single latent space of outfit codes and a single draping network. This is made possible because of the choice of the point cloud representation and the use of topology-independent, point cloud-specific losses during the learning of the joint model. After learning, the model is capable of generalizing to new outfits, capturing their geometry from data, and to drape the acquired outfits over bodies of varying shapes and in new poses. With our model, acquiring the outfit geometry can be done from as little as a single image.

We extend our approach beyond geometry acquisition to include the appearance modeling. Here, we use the ideas of differentiable rendering~\cite{opendr,ravi2020pytorch3d,nvdiffrast} and neural point-based graphics~\cite{npbg,rerendering,wiles2020synsin}. Given a video sequence of an outfit worn by a person, we capture the photometric properties of the outfit using neural descriptors attached to points in the point cloud, and the parameters of a rendering (decoder) network. The fitting of the neural point descriptors and the rendering network (which capture the photometric properties) is performed jointly with the estimation of the outfit code (which captures the outfit geometry) within the same optimization process. After fitting, the outfit can be transferred and re-rendered in a realistic way over new bodies and in new poses.

In the experiments, we evaluate the ability of our geometric model to capture the deformable geometry of new outfits using point clouds. We further test the capability of our full approach to capture both outfit geometry and appearance from videos and to re-render the learned outfits to new targets. The experimental comparisons show the viability of the point-based approach to clothing modeling. We will publish our code and model at \url{https://saic-violet.github.io/point-based-clothing/}.

\section{Related work on clothing modeling}

\paragraph{Modeling clothing geometry.}  Many existing methods model clothing geometry using one or several pre-defined garment templates of fixed topology. DRAPE \cite{DRAPE2012}, which is one of the earlier works, learns from Physics-based simulation (PBS) and allows for pose and shape variation for each learned garment mesh. Newer works usually represent garment templates in the form of offsets (displacements) to SMPL~\cite{SMPL:2015} mesh. ClothCap \cite{Moll17clothcap} employs such a technique and captures more fine-grained details learned from the new dataset of 4D scans. DeepWrinkles \cite{laehner2018deepwrinkles} also addresses the problem of fine-grained wrinkles modeling with the use of normal maps generated by a conditional GAN. GarNet \cite{gundogdu2019garnet} incorporates two-stream architecture and makes it possible to simulate garment meshes at the level of realism that almost matches PBS, while being two orders of magnitude faster. TailorNet \cite{patel2020tailornet} follows the same SMPL-based template approach as \cite{Moll17clothcap,bhatnagar2019mgn} but models the garment deformations as a function of pose, shape and style simultaneously (unlike the previous work). It also shows greater inference speed than \cite{gundogdu2019garnet}. The CAPE system~ \cite{CAPE:CVPR:20} uses graph ConvNet-based generative shape model that enables to condition, sample, and preserve fine shape detail in 3D meshes.

Several other works recover clothing geometry simultaneously with the full body mesh from image data. BodyNet \cite{varol18_bodynet} and DeepHuman \cite{Zheng2019DeepHuman} are voxel-based methods that directly infer the volumetric dressed body shape from a single image. In SiCloPe \cite{natsume2019siclope} the authors use similar approach, but synthesize the silhouettes of the subjects in order to recover more details. HMR \cite{kanazawa2018endtoend} utilizes SMPL body model to estimate pose and shape from an input image. Some approaches such as PIFu \cite{saito2019pifu} and ARCH \cite{huang2020arch} employ end-to-end implicit functions for clothed human 3D reconstruction and are able to generalise to complex clothing and hair topology, while PIFuHD \cite{saito2020pifuhd} recovers higher resolution 3D surface by using two-level architecture. However, these SDF approaches can only represent closed connected surfaces, whereas point clouds may represent arbitrary topologies. 

MouldingHumans \cite{gabeur2019moulding} predicts the final surface from estimated ``visible'' and ``hidden'' depth maps. MonoClothCap \cite{xiang2020monoclothcap} demonstrates promising results in video-based temporally coherent dynamic clothing deformation modeling. Most recently, Yoon~et~al.~\cite{yoon2021neural} design relatively simple yet effective pipeline for template-based garment mesh retargeting.

Our geometric modeling differs from previous works through the use of a different representations (point clouds), which gives our approach topological flexibility, the ability to model clothing separately from the body, while also providing the geometric scaffold for appearance modeling with neural rendering.

\vspace{-12pt}

\paragraph{Modeling clothing appearance.}  A large number of work focus on direct image-to-image transfer of clothing bypassing 3D modeling. Thus, \cite{jetchev2017conditional,han2018viton,wang2018characteristicpreserving,yang2020photorealistic,issenhuth2020mask} address the task of transferring a desired clothing item onto the corresponding region of a person given their images. CAGAN \cite{jetchev2017conditional} is one of the first works that proposed to utilize image-to-image conditional GAN to tackle this task. VITON \cite{han2018viton} follows the idea of image generation and uses non-parametric geometric transform which makes all the procedure two-stage, similar to SwapNet \cite{raj2018swapnet} with the difference in the task statement and training data. CP-VTON \cite{wang2018characteristicpreserving} further improves upon \cite{han2018viton} by incorporating a full learnable thin-plate spline transformation, followed by CP-VTON+ \cite{Minar_CPP_2020_CVPR_Workshops}, LA-VITON \cite{Lee_2019_ICCV}, Ayush~et~al.~\cite{Ayush_2019_ICCV} and ACGPN~\cite{yang2020photorealistic}. While the above-mentioned works rely on pre-trained human parsers and pose estimators, the recent work of Issenhuth~er~al.~\cite{issenhuth2020mask} achieves competitive image quality and significant speed-up by employing a teacher-student setting to distill the standard virtual try-on pipeline. The resulting student network does not invoke an expensive human parsing network at inference time. Very recently introduced VOGUE~\cite{lewis2021vogue} train a pose-conditioned StyleGAN2~\cite{karras2020analyzing} and find the optimal combination of latent codes to produce high-quality try-on images. 

Some methods make use of both 2D and 3D information for model training and inference. Cloth-VTON \cite{Minar_2020_ACCV} employs 3D-based warping to realistically retarget a 2D clothing template. Pix2Surf \cite{mir20pix2surf} allows to digitally map the texture of online retail store clothing images to the 3D surface of virtual garment items enabling 3D virtual try-on in real-time. Other relevant research extend the scenario of single template cloth retargeting to multi-garment dressing with unpaired data \cite{Neuberger_2020_CVPR}, generating high-resolution fashion model images wearing custom outfits \cite{Yildirim_2019_ICCV}, or editing the style of a person in the input image \cite{hsiao2019fashion}. 

In contrast to the referenced approaches to clothing appearance retargeting, ours uses explicit 3D geometric models, while not relying on individual templates of fixed topology. On the downside, our appearance modeling part requires video sequence, while some of the referenced works use one or few images.

\paragraph{Joint modeling of geometry and appearance.} Octopus~\cite{alldieck2019learning} and Multi-Garment Net (MGN)~\cite{bhatnagar2019mgn} recover the textured clothed body mesh based on the SMPL+D model. The latter method treats clothing meshes separately from the body mesh, which gives it the ability to transfer the outfit to another subject. Tex2Shape \cite{alldieck2019tex2shape} proposes an interesting framework that turns the shape regression task into an image-to-image translation problem. In \cite{shen2020garmentgeneration}, a learning-based parametric generative model is introduced that can support any type of garment material, body shape, and most garment topologies. Very recently, StylePeople~\cite{Iskakov21} approach integrates polygonal body mesh modeling with neural rendering, so that both clothing geometry and the texture are encoded in the neural texture~\cite{thies2019deferred}. Similarly to \cite{Iskakov21} our approach to appearance modeling also relies on neural rendering, however our handling of geometry is more explicit. In the experiments, we compare to \cite{Iskakov21} and observe the advantage of a more explicit geometric modeling, especially for loose clothing.

Finally, we note that in parallel with us, the SCALE system~\cite{Ma:CVPR:2021} explored very similar ideas (point-based geometry modeling and its combination with neural rendering) for modeling clothed humans from 3D scans.

\section{Method}
We first discuss the point cloud draping model. The goal of this model is to capture the geometry of diverse human outfits draped over human bodies with diverse shapes and poses using point clouds. We propose a latent model for such point clouds that can be fitted to a single image or to more complete data. We then describe the combination of the point cloud draping with neural rendering that allows us to capture the appearance of outfits from videos.

\subsection{Point cloud draping}

\paragraph{Learning the model.} 
\label{glo_intro}
We learn the model using generative latent optimization (GLO) \cite{bojanowski2019optimizing}. We assume that the training set has a set of $N$ outfits, and associate each outfit with $d$-dimensional vector $z$ (the $\textit{outfit code}$). We thus randomly initialize $\{z_1, \ldots , z_N\}$, where $z_i \in Z \subseteq \mathbb{R}^d$ for all $i = 1, \ldots , N$.

During training, for each outfit, we observe its shape for a diverse set of human poses. The target shapes are given by a set of geometries.  In our case, we use synthetic CLOTH3D dataset~\cite{bertiche2020cloth3d} that provides shapes in the form of meshes of varying topology. In this dataset, each subject is wearing an outfit and performs a sequence of movements. For each outfit $i$ for each frame $j$ in the corresponding sequence, we sample points from the mesh of this outfit and obtain the point cloud $x_i^j \in X$, where $X$ denotes the space of point clouds of a fixed size ($8192$ is used in our experiments). We denote the length of the training sequence of the $i$-th outfit as $P_i$. We also assume that the body mesh $s_i^j \in S$ is given, and in our experiments we work with the SMPL~\cite{smpl} mesh format (thus $S$ denotes the space of SMPL meshes for varying body shape parameters and body pose parameters). Putting it all together, we obtain the dataset $\{(z_i, s_i^j, x_i^j)\}_{i=1.. N,\,j=1..P_i}$ of outfit codes, SMPL meshes, and clothing point clouds.

As our goal is to learn to predict the geometry in new poses and for new body shapes, we introduce the \textit{draping function} $G_{\theta}: Z \times S \to X$ that maps the latent code and the SMPL mesh (representing the naked body) to the outfit point cloud. Here, $\theta$ denotes the learnable parameters of the function. We then perform learning by the optimization of the following objective:
\begin{equation} \label{eq:glo}
\min_{\theta \in \Theta \atop \{z_1,\ldots{},z_N\}}{ \frac{1}{N} \sum_{i=1}^{N} \frac{1}{P_i} \sum_{j=1}^{P_i} L_{\text{3D}}\left(G_{\theta}(z_i, s_i^j),\, x_i^j\right)}
\end{equation}
In \eq{glo}, the objective is the mean reconstruction loss for the training point clouds over the training set. The loss $L_{\text{3D}}$ is thus the 3D reconstruction loss. In our experiments, we use the approximate Earth Mover's Distance~\cite{liu2019morphing}. Note, that as this loss measures the distance between point clouds and ignores all topological properties, our learning formulation is naturally suitable for learning outfits of diverse topology.

We perform optimization jointly over the parameters of our draping function $G_{\theta}$ and over the latent outfit code $z_i$ for all $i = 1, \ldots , N$. Following \cite{bojanowski2019optimizing}, to regularize the process, we clip the outfit codes to the unit ball during optimization. The optimization process thus establishes the outfit latent code space and the parameters of the draping function.

\paragraph{Draping network.} We implement the draping function $G_{\theta}(z,s)$ as a neural network that takes the SMPL mesh $s$ and transforms this point cloud into the outfit point cloud. Over the last years, point clouds have become (almost) first-class citizens in the deep learning world, as a number of architectures that can input and/or output point clouds and operate on them have been proposed. In our work, we use the recently introduced \textit{Cloud Transformer} architecture~\cite{mazur2020cloud} due to its capability to handle diverse point cloud processing tasks. 

The cloud transformer comprises of blocks, each of which sequentially rasterizes, convolves, and de-rasterizes the point cloud at the learned data-dependent positions. The cloud transformer thus deforms the input point cloud (derived from the SMPL mesh as discussed below) into the output point cloud $x$ over a number of blocks. We use a simplified version of the cloud transformer with single-headed blocks to reduce the  computational complexity and memory requirements. Otherwise, we follow the architecture of the generator suggested in \cite{mazur2020cloud} for image-based shape reconstruction, which in their case takes the point cloud (sampled from the unit sphere) and a vector (computed by the image encoding network) as an input and outputs the point cloud of the shape depicted in the image.

\begin{figure}
  \includegraphics[width=\linewidth]{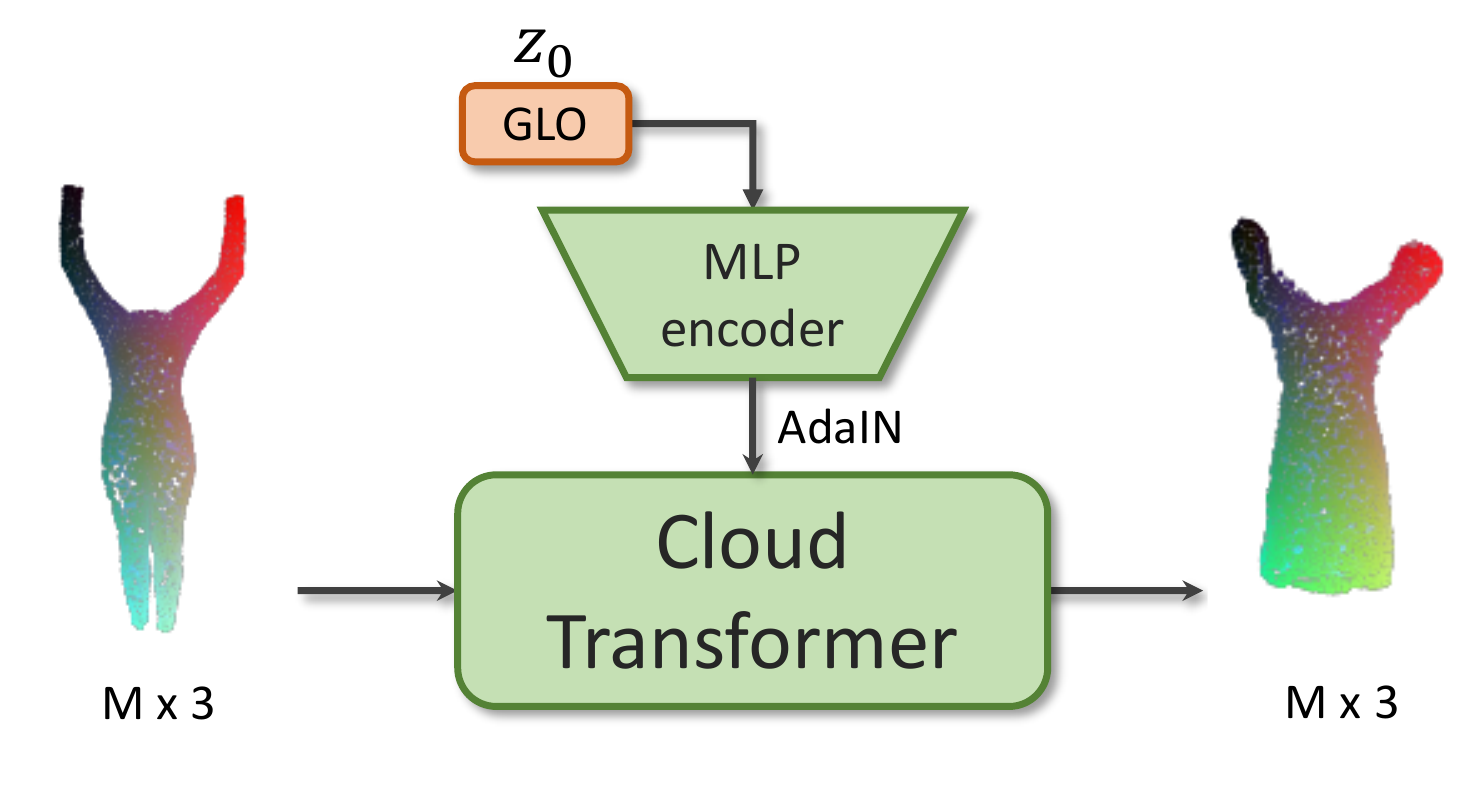}
  \caption{Our draping networks morphs the body point cloud (left) and the outfit code (top) into the outfit point cloud that is adapted to the body pose and the body shape.}
  \vspace{-10pt}
\end{figure}

In our case, the input point cloud and the vector are different and correspond to the SMPL mesh and the outfit code respectively. More specifically, to input the SMPL mesh into the cloud transformer architecture, we first remove the parts of the mesh corresponding to the head, the feet and the hands. We then consider the remaining vertices as a point cloud. To densify this point cloud, we also add the midpoints of the SMPL mesh edges to this point cloud. The resulting point cloud (which is shaped by the SMPL mesh and reflects the change of pose and shape) is input into the cloud transformer.

Following~\cite{mazur2020cloud}, the latent outfit code $z$ is input into the cloud transformer through AdaIn connections~\cite{huang2017arbitrary} that modulate the convolutional maps inside the rasterization-derasterization blocks. The particular weights and biases for each AdaIn connection are predicted from the latent code $z$ via a perceptron, as is common for style-based generators~\cite{stylegan}. We note that while we have obtained good results using the (simplified) cloud transformer architecture, other deep learning architectures that operate on point clouds (e.g.~PointNet~\cite{pointnet}) can be employed.

\begin{figure}
    \centering
    \includegraphics[width=\columnwidth]{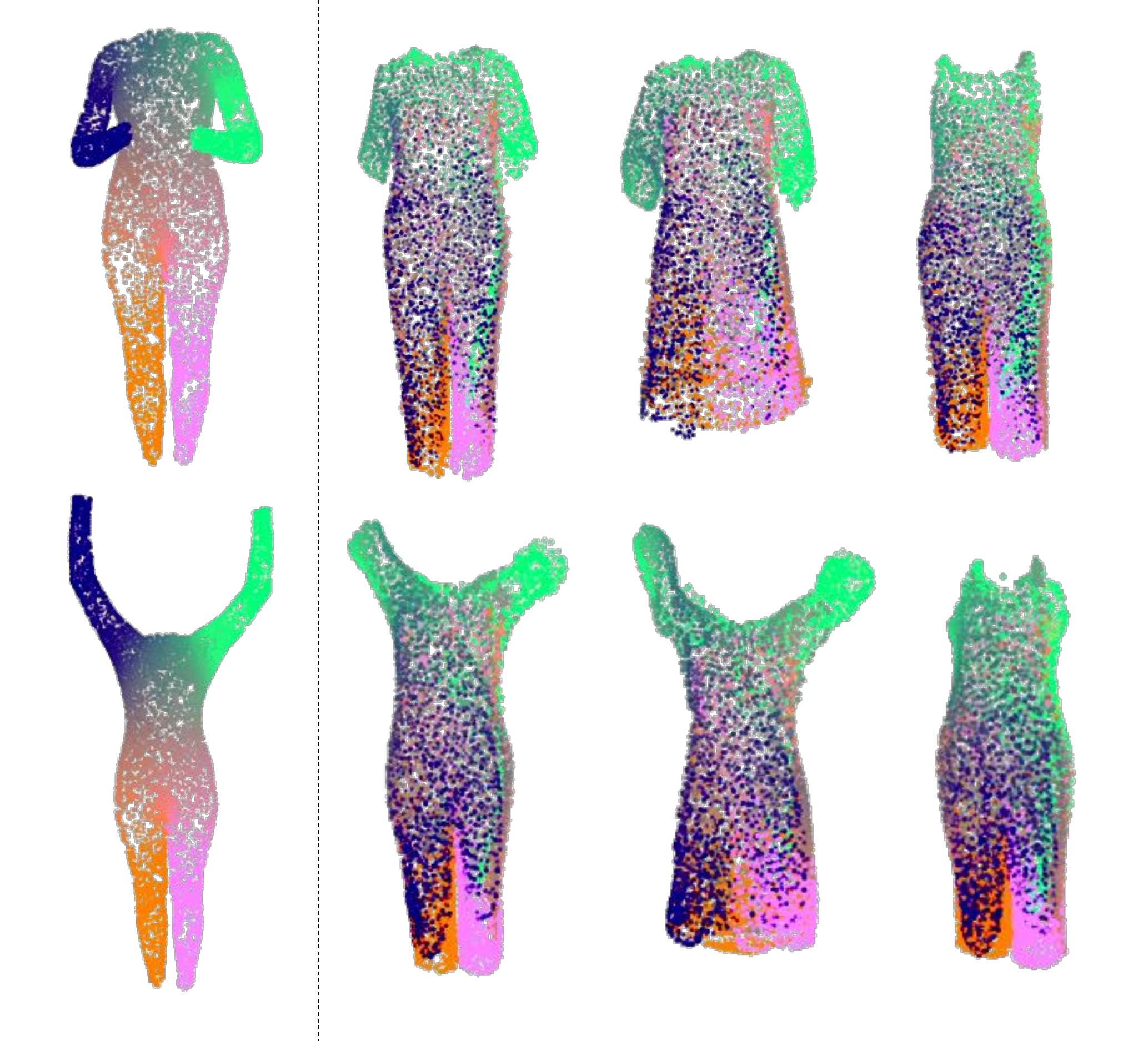}
    \caption{More color-coded results of the draping networks. Each row corresponds to a pose. The leftmost image shows the input to the draping network. The remaining columns correspond to three outfit codes. Color coding corresponds to spectral coordinates on the SMPL mesh surface. Color coding reveals that the draping transformation is noticeably non-local (i.e.~the draping network does not simply compute local displacements). Also, color coding reveals correspondences between analogous parts of outfit point clouds across the draping network outputs.}
    \vspace{-10pt}
    \label{fig:folding}
\end{figure}

We also note that the morphing implemented by the draping network is strongly non-local (i.e.~our model does not simply compute local vertex displacements), and is consistent across outfits and poses (\fig{folding}).

\paragraph{Estimating the outfit code.} 
\label{style_search}
Once the draping network is pre-trained on a large synthetic dataset~\cite{bertiche2020cloth3d}, we are able to model the geometry of a previously unseen outfit. The fitting can be done for a single or multiple images. For a single image, we optimize the outfit code $z^*$ to match the segmentation mask of the outfit in the image.

In more detail, we predict the binary outfit mask by passing given RGB image through Graphonomy network~\cite{Gong2019Graphonomy} and combining all semantic masks that correspond to clothing. We also fit the SMPL mesh to the person in the image using the SMPLify approach~\cite{simplify}. We then minimize the 2D Chamfer loss between the outfit segmentation mask and the projection of the predicted point-cloud onto the image. The projection takes into account the occlusions of the outfit by the SMPL mesh (e.g.~the back part of the outfit when seen from the front). In this case, the optimization is performed over the outfit code $z^*$ while the parameters of the draping network remain fixed to avoid overfitting to a single image. 

For complex outfits we observed instability in the optimization process, which often results in undesired local minima. To find a better optimum, we start from several random initializations $\{ z^*_1, \ldots z^*_T \}$ independently (in our experiments, $T = 4$ random initializations are used). After several optimization steps we take the average outfit vector $\bar{z} = \frac{1}{T} \sum_{t=1}^T z^{*}_t$ and then continue the optimization from $\bar{z}$ until convergence. We observed that this simple technique provides consistently accurate outfit codes. Typically we make $100$~training steps while optimizing $T$ hypothesis. After the averaging the optimization takes $50-400$~steps depending on the complexity of the outfit's geometry. 

\subsection{Appearance modeling}

\begin{figure}
  \includegraphics[width=\linewidth]{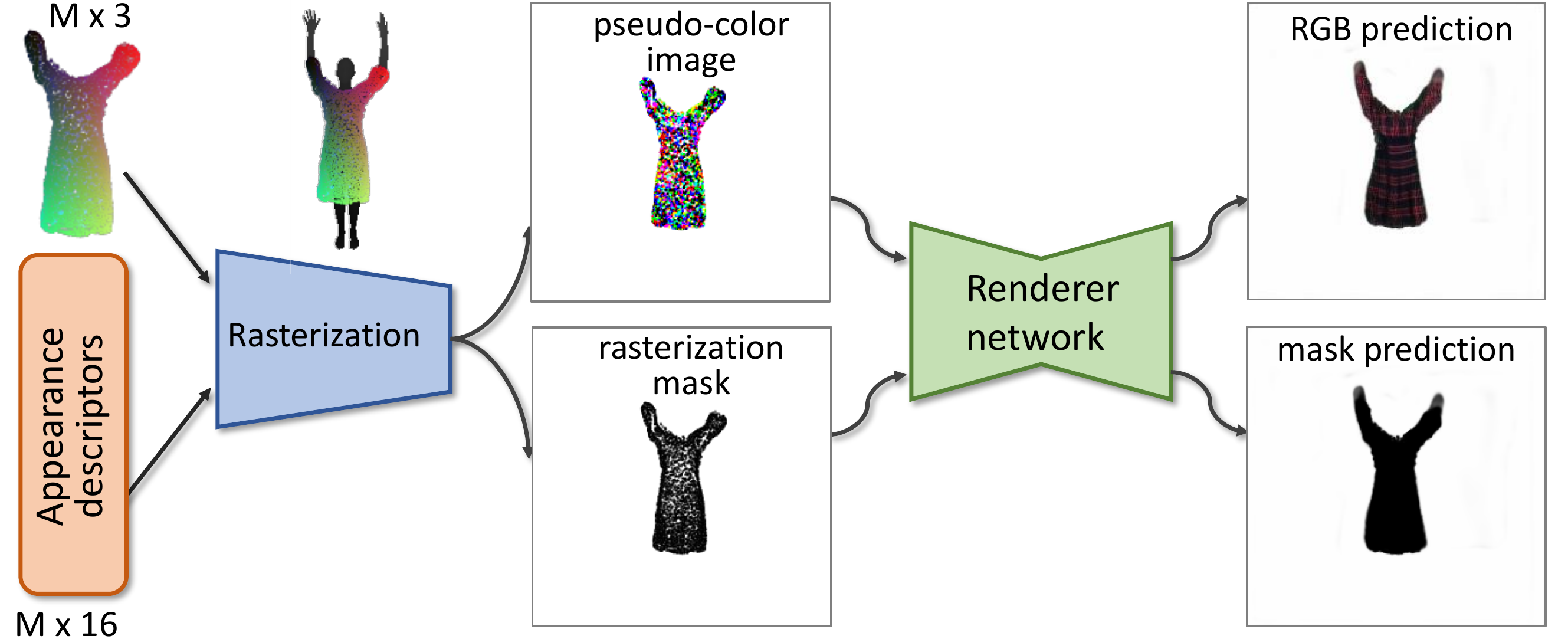}
  \caption{We use neural point-based graphics to model the appearance of an outfit. We thus learn the set of neural appearance descriptors and the renderer network that allow to translate the rasterization of the outfit point cloud into its realistic masked image (right).}
  \label{fig:appearance}
  \vspace{-10pt}
\end{figure}

\paragraph{Point-based rendering.} Most applications of clothing modeling go beyond geometric modeling and require to model the appearance as well. Recently, it has been shown that point clouds provide good geometric scaffolds for neural rendering~\cite{npbg,wiles2020synsin,rerendering}. We follow the neural point-based graphics (NPBG) modeling approach~\cite{npbg} to add appearance modeling to our system (\fig{appearance}).

Thus, when modeling the appearance of a certain outfit with the outfit code $z$, we attach $p$-dimensional latent appearance vectors $T=\{t[1],\ldots{},t[M]\}$ to each of the $M$ points in the point cloud that models its geometry. We also introduce the rendering network $R_\psi$ with learnable parameters $\psi$. To obtain the realistic rendering of the outfit given the body pose $s$ and the camera pose $C$, we then first compute the point cloud $G_\theta(z,s)$, and then rasterize the point cloud over the image grid of resolution $W\times{}H$ using the camera parameters and the neural descriptor $t[m]$ as a pseudo-color of the $m$-th point. We concatenate the result of the rasterization, which is a $p$-channeled image, with the rasterization masks, which indicates non-zero pixels, and then process (translate) them into the outfit RGB color image and the outfit mask (i.e.~a four-channel image) using the rendering network $R_\psi$ with learnable parameters $\psi$. 

During the rasterization, we also take into account the SMPL mesh of the body and do not rasterize the points occluded by the body. For the rendering network we use a lightweight U-net network~\cite{unet}. 

\vspace{-14pt}
\paragraph{Video-based appearance capture.} Our approach allows to capture the appearance of the outfit from video. To do that we perform two-stage optimization. In the first stage, the outfit code is optimized, minimizing the Chamfer loss between the point cloud projections and the segmentation masks, as described in the previous section. Then, we jointly optimize latent appearance vectors $T$, and the parameters of the rendering network $\psi$. For the second stage we use (1) the perceptual loss~\cite{perceptual} between the masked video frame and the RGB image rendered by our model, and (2) the Dice loss between the segmentation mask and the rendering mask predicted by the rendering network.

Appearance optimization requires a video of a a person with whole surface of their body visible in at least one frame. In our experiments training sequences consist of 600 to 2800 frames for each person. The whole process takes roughly 10 hours on NVIDIA Tesla P40 GPU.

After the optimization, the acquired outfit model can be rendered for arbitrarily posed SMPL body shapes, providing RGB images and segmentation masks.

\section{Experiments}

We evaluate the geometric modeling and the appearance modeling within our approach and compare it to prior art. Please also refer to the supplementary video on the \fnurl{project page}{https://saic-violet.github.io/point-based-clothing} for a more convenient demonstration of qualitative comparison.
\vspace{-14pt}
\paragraph{Datasets.} We use the Cloth3D~\cite{bertiche2020cloth3d} dataset to train our geometric meta-model. The Cloth3D dataset has 11.3K garment elements of diverse geometry modeled as meshes draped over 8.5K SMPL bodies undergoing pose changes. The fitting uses physics-based simulation. We split the Cloth3D dataset into 6475 training sequences and 1256 holdout sequences, where sequences differ by SMPL parameters and outfit mesh. 

We evaluate both stages - geometry and appearance -  using two datasets of human videos. These datasets do not contain 3D data and were not used during the draping network training. The \textit{PeopleSnapsot} presented in~\cite{alldieck2018video} contains 24 videos of people in diverse clothes rotating in A-pose. In terms of clothing, it lacks examples of people wearing skirts and thus does not reveal the full advantage of our method. We also evaluate on a subset from \textit{AzurePeople} dataset introduced in~\cite{Iskakov21}. This subset contains videos of eight people in outfits of diverse complexity shot from 5 RGBD Kinect cameras. For both datasets we generate cloth segmentation masks with Graphonomy method~\cite{Gong2019Graphonomy} and SMPL meshes using SMPLify~\cite{simplify}. To run all approaches in our comparison, we also predict Openpose~\cite{cao2019openpose} keypoints, DensePose~\cite{guler2018densepose} UV renders and SMPL-X~\cite{pavlakos2019expressive} meshes. For appearance modeling, we follow StylePeople’s procedure and use the data from four cameras as a training set and validate by the fifth (the leftmost) camera. 

We note that the two evaluation datasets (PeopleSnapshot and AzurePeople) were not seen during the training of the draping network. Furthermore, the comparisons in this section and all the visualizations in the supplementary material are obtained given the previously unseen outfit segmentations. The poses and body shapes were also sampled from the holdout set and were not seen by the draping network and by the rendering network during their training. By this, we emphasize the ability of our approach to generalize to new outfit styles, new body poses, and new body shapes.

\subsection{Details of the draping network}
To build a geometric prior on clothing, our draping function $G_{\theta}$ is pre-trained on synthetic Cloth3D dataset. We split it into train and validation parts, resulting in $N=6475$ training video sequences. Since most of the consequent frames share similar pose/clothes geometry, only every tenth frame is considered for the training. As described in Sec.~\ref{glo_intro}, we randomly initialize $\{z_1, \ldots , z_N\}$, where $z_i \in Z \subseteq \mathbb{R}^d$ for each identity $i$ in the dataset. In our experiments, we set the latent code dimensionality relatively low to $d{=}8$, in order to avoid overfitting during subsequent single-image shape fitting (as described in~Sec.~\ref{style_search}). 

We feed the outfit codes $z_i$ to an MLP encoder consisting of $5$ fully-connected layers to obtain a 512-dimensional latent representation. Then it is passed to the AdaIn branch of the Cloud Transformer network. For pose and body information, we feed an SMPL point cloud with hands, feet and head vertices removed, see \fig{garment}. The draping network outputs three-dimensional point clouds with $8.192$ points in all experiments.  We choose approximate Earth Mover's Distance~\cite{liu2019morphing} as the loss function and optimize each GLO-vector and the draping network simultaneously using Adam~\cite{kingma2017adam}. 

While our pre-traning provides expressive priors on dresses and skirts, the ability of the model to produce tighter outfits is somewhat limited. We speculate that this effect is mainly caused by a high bias towards jumpsuits in the Cloth3D tight clothing categories.

\begin{table}
\centering
\resizebox{0.46\textwidth}{!}{
 \begin{tabular}{c c c c} 
 
  & Ours v Tex2Shape & Ours v MGN & Ours v Octopus \\
 \hline
 \textit{PeopleSnapshot} & 38.1\% vs 61.9\% & 50.9\% vs 49.1\% & 47.8\% vs 52.2\%  \\
 \textit{AzurePeople} & 65.6\% vs 34.3\% &  74.5\% vs 25.5\% &  73.7\% vs 26.3\% \\
\end{tabular}
}
\caption{Results of user study, in which the users compared the quality of 3D clothing geometry recovery (fitted to a single image). Our method is preferred on the AzurePeople dataset with looser clothing, while the previously proposed methods work better for tighter clothing of fixed topology. 
\vspace{-10pt}}
\label{tab:userstudy}
\end{table}

\subsection{Recovering outfit geometry}

\begin{figure}
  \includegraphics[width=\columnwidth]{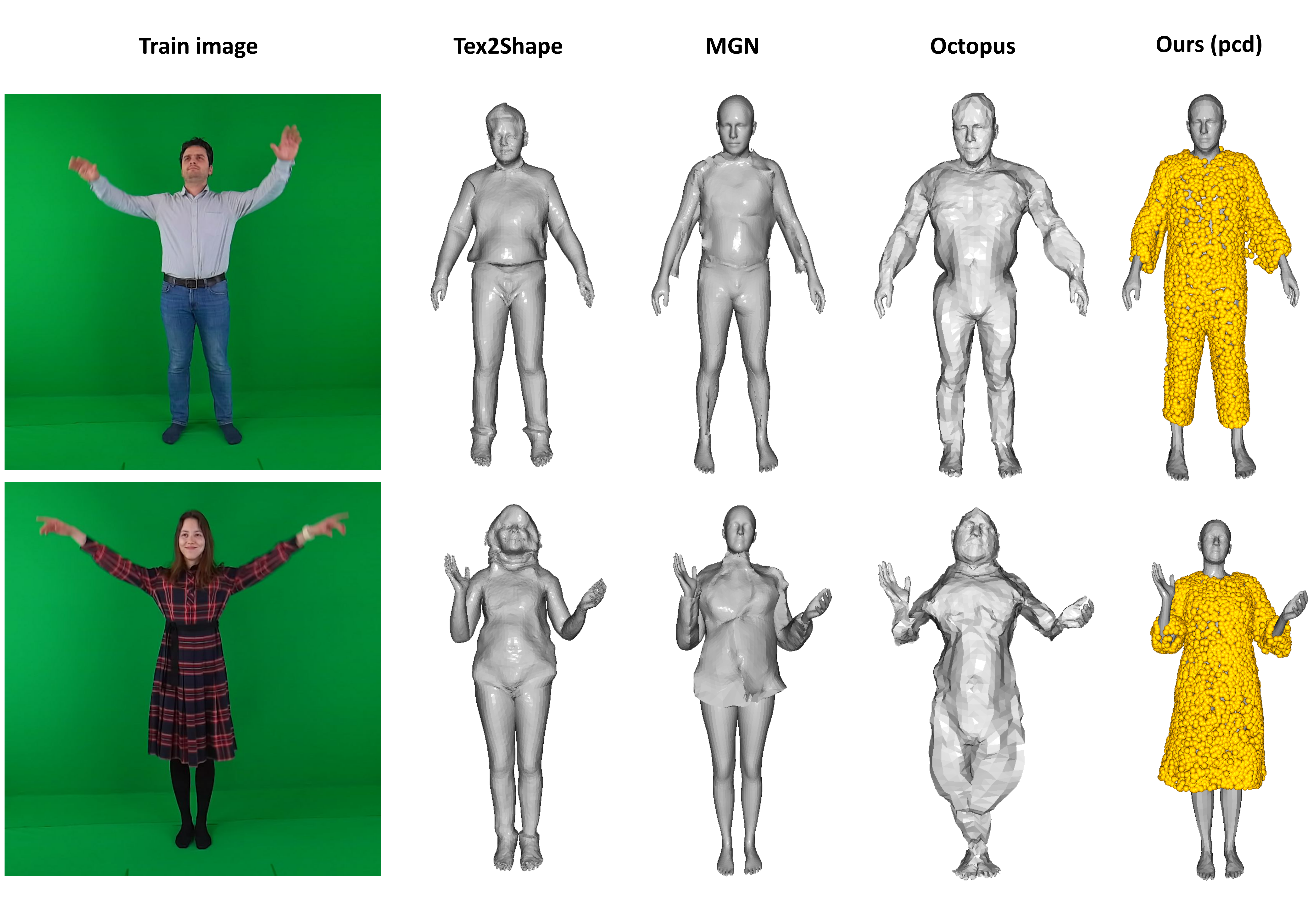}
  \caption{We show the predicted geometries in the validation poses fitted to a single frame (left). 
  For our method (right) the geometry is defined by a point cloud (shown in yellow), while for Tex2Shape and MultiGarmentNet (MGN) the outputs are mesh based. Our method is able to reconstruct the dress, while other methods fail (bottom row). Note, our method is able to reconstruct a tighter outfit too (top row), though Tex2Shape with its displacement-based approach achieves a better result in this case.}
  \vspace{-14pt}
  \label{fig:geomcomparison}
\end{figure}

In this series of experiments, we evaluate the ability of our method to recover the outfit geometry from a single photograph. We compare \textit{Ours} point-based approach with the following three methods:
\begin{enumerate}
    \setlength\itemsep{0.2em}
    \item The \textit{Tex2Shape} method~\cite{alldieck2019tex2shape} that predicts offsets for vertices of SMPL mesh in texture space. It is ideally suited for the PeopleSnapshot dataset, while less suitable to AzurePeople sequences with skirts and dresses.
    
    \item The \textit{Octopus} work~\cite{alldieck2019learning} uses the displacements to the SMPL body model vertices to reconstruct a full-body human avatar with hair and clothing. Though the authors note that it is not ideally suited for reconstruction based on a single photograph. 
    
    \item The \textit{Multi-garment net} approach~\cite{bhatnagar2019mgn} builds upon Octopus and predicts upper and lower clothing as separate meshes. It proposes a virtual wardrobe of pre-fitted garments, and is also able to fit new outfits from a single image.
\end{enumerate}
We note that the compared systems use different formats to recover clothing (point cloud, vertex offsets, meshes). Furthermore, they are actually solving slightly different problems, as our method and Multi-garment net recover clothing, while Tex2Shape recovers meshes that comprise clothing, body, and hair. All three systems, however, support retargeting to new poses. We therefore decided to evaluate the relative performance of the three methods through a user study that assesses the realism of clothing retargeting.

% To make our method more comparable to the competitors, we convert the recovered point clouds into meshes using the approach~\cite{kazdan2006possion} after retargeting.
We present the users with triplets of images, where the middle image shows the source photograph, while the side images show the results of two compared methods (in the form of shaded mesh renders for the same new pose). The result of such pairwise comparisons (user preferences) aggregated over $\sim$1.5k user comparisons are shown in \tab{userstudy}. Our method is strongly preferred by the users in the case of AzurePeople dataset that contains skirts and dresses, while Tex2Shape and MGN are prefered on PeopleSnapshot dataset that has tighter clothing with fixed topology. \fig{geomcomparison} shows typical cases, while the supplementary material provides more extensive qualitative comparisons. Note, in user study we paint our points in gray to exclude the coloring factor in user's choice. 

Since our approach uses 2D information to fit outfit code, we decided to omit quantitative comparison by the standard metrics due to the lack of datasets that contain both realistic RGB and realistic 3D data. However, we compare our method to MGN on BCNet~\cite{jiang2020bcnet} dataset. For both methods, we use projection masks of the outfit meshes to fit clothing geometry. Our approach fits clothing geometry better in terms of Chamfer distance to vertices of ground  truth outfit meshes in validation poses ($0.00121$ vs $0.0025$) on $200$ randomly chosen samples.

\begin{figure}[h]
    \centering
    % trim={<left> <lower> <right> <upper>}
    \includegraphics[width=0.5\textwidth,trim={0 2cm 0 2cm},clip]{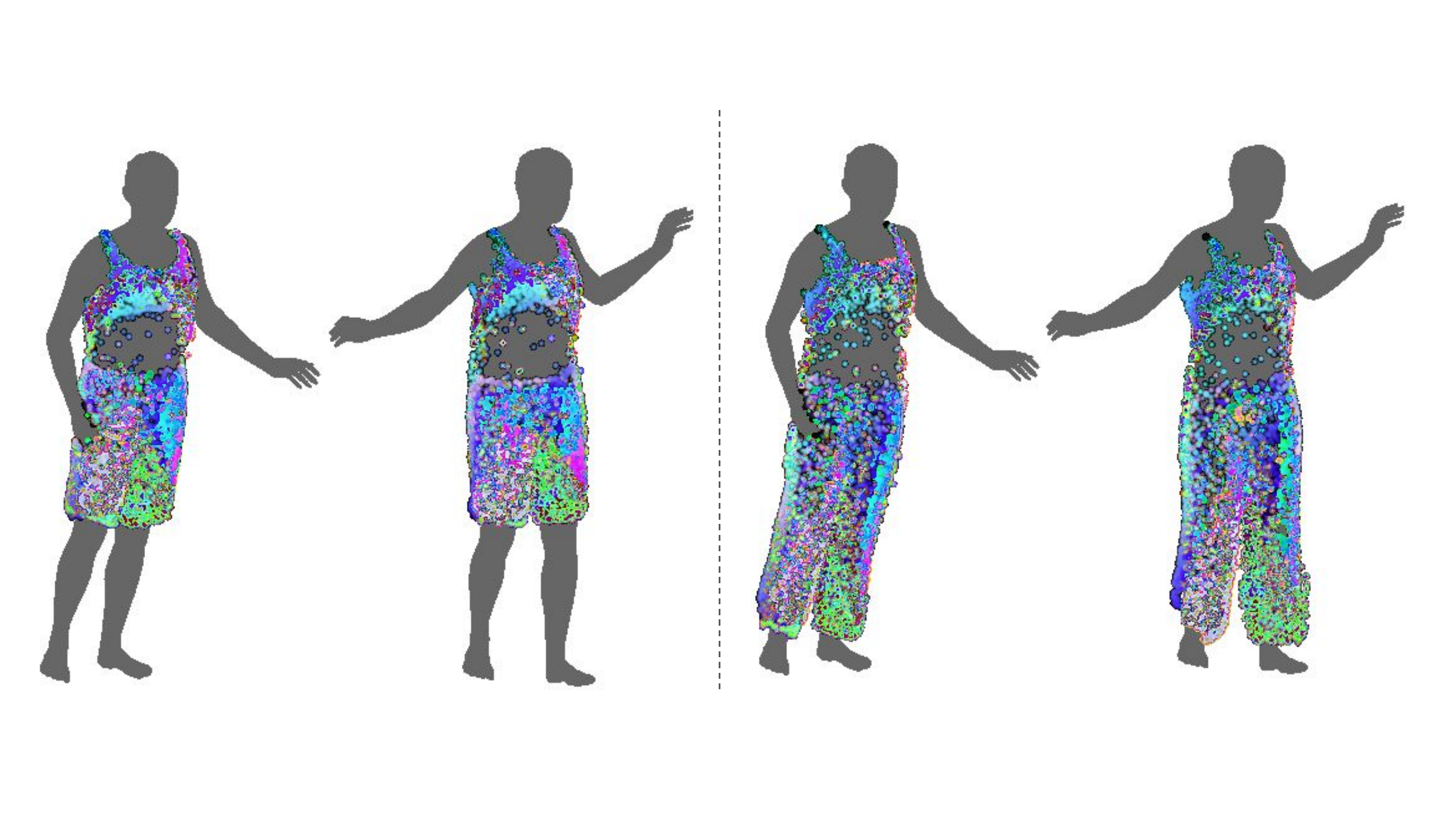}
    \caption{Our method is also capable of modeling the separated top and bottom garment styles. Here two different outfits in two different poses are shown.}
    \label{fig:separable_graments}
\end{figure}

\vspace{-15pt}
\subsection{Appearance modeling}

\begin{figure*}[h!]
    \setlength{\tabcolsep}{.1em}
    \centering
    \vspace{0.cm}
    \newcommand{\rdrs}[2]{
            \includegraphics[trim={#2 0 #2 0},clip,height=2.8cm]{images/rgb_comp/#1_gt.png} &
            \includegraphics[trim={#2 0 #2 0},clip,height=2.8cm]{images/rgb_comp/#1_sp.png} &
            \includegraphics[trim={#2 0 #2 0},clip,height=2.8cm]{images/rgb_comp/#1_ours.png}  
        }

    \centering
    \begin{tabular}{ccccccccc}
         \rdrs{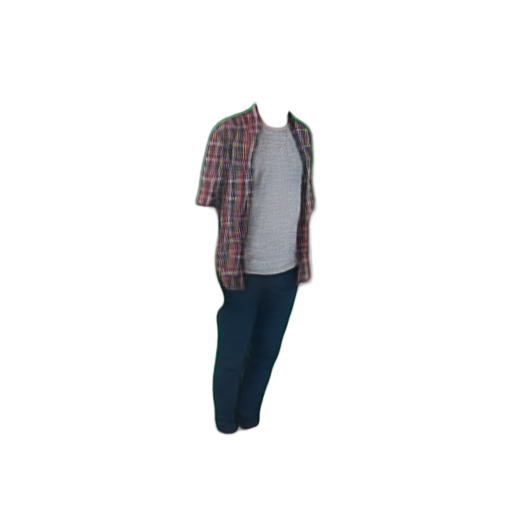}{100} \,\,& \rdrs{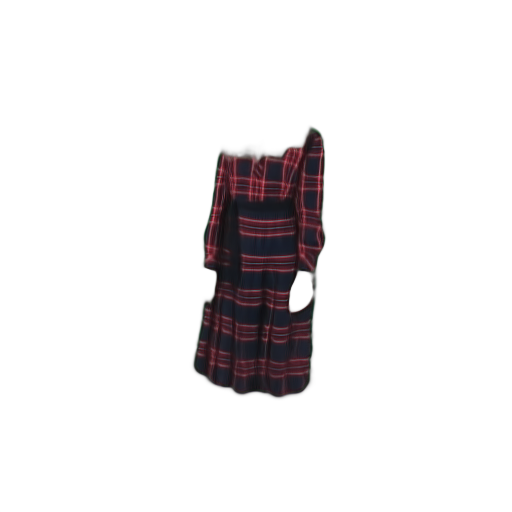}{100} \,\,& \rdrs{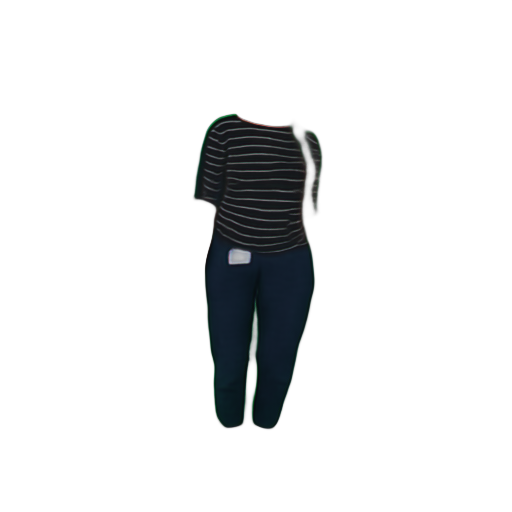}{100} \\
         \rdrs{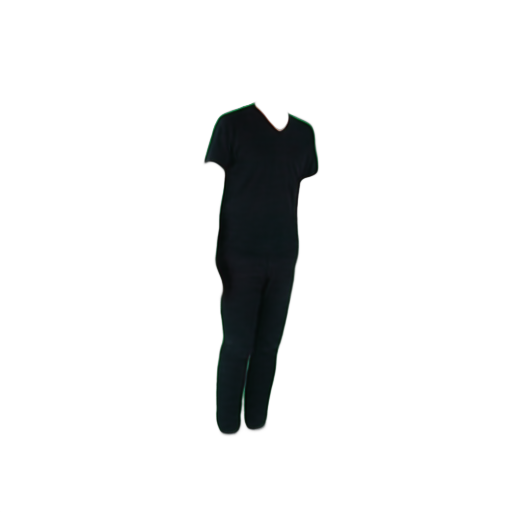}{100} \,\,& \rdrs{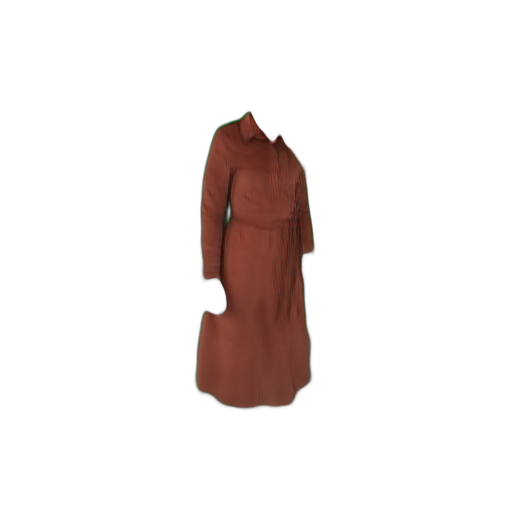}{100} \,\,& \rdrs{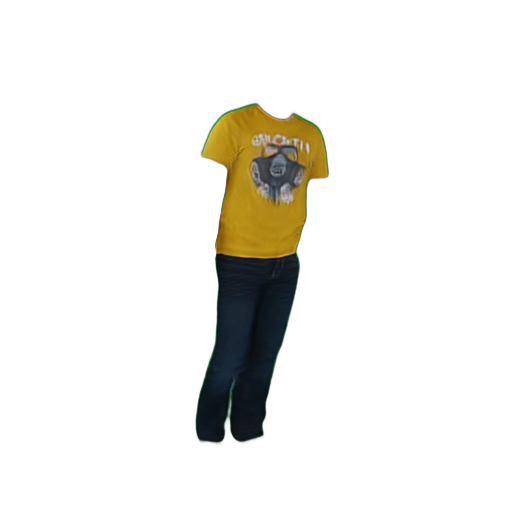}{100} \\
         Input & StylePeople & Ours &
         Input & StylePeople & Ours &
         Input & StylePeople & Ours 
    \end{tabular}
    \caption{We compare the appearance retargeting results of our method to new poses unseen during fitting between our method and the StylePeople system (multi-shot variant), which uses the SMPL mesh as the underlying geometry and relies on neural rendering alone to ``grow'' loose clothes in renders. As expected, our system produces sharper results for looser clothes due to the use of more accurate geometric scaffolding. \textit{Zoom-in is highly recommended.}}
    \label{fig:rgbcomp}
\end{figure*}

We evaluate our appearance modeling pipeline against the \textit{StylePeople} system~\cite{Iskakov21} (the multiframe variant) that is the closest to ours in many ways. StylePeople fits a neural texture of the SMPL-X mesh alongside the rendering network using a video of a person using backpropagation. For comparison purposes we modify StylePeople to generate clothing masks along with rgb images and foreground segmentations. Both approaches are trained separately on each person from \textit{AzurePeople} and \textit{PeopleSnapshot} dataset. We then compare outfit images generated for holdout views in terms of three metrics that measure visual similarity to ground truth images, namely learned perceptual similarity (\textit{LPIPS})~\cite{zhang2018unreasonable} distance, structural similarity (\textit{SSIM})~\cite{Wang04} and its multiscale version (\textit{MS-SSIM}). 

\begin{figure}
    \centering
    \includegraphics[width=\columnwidth]{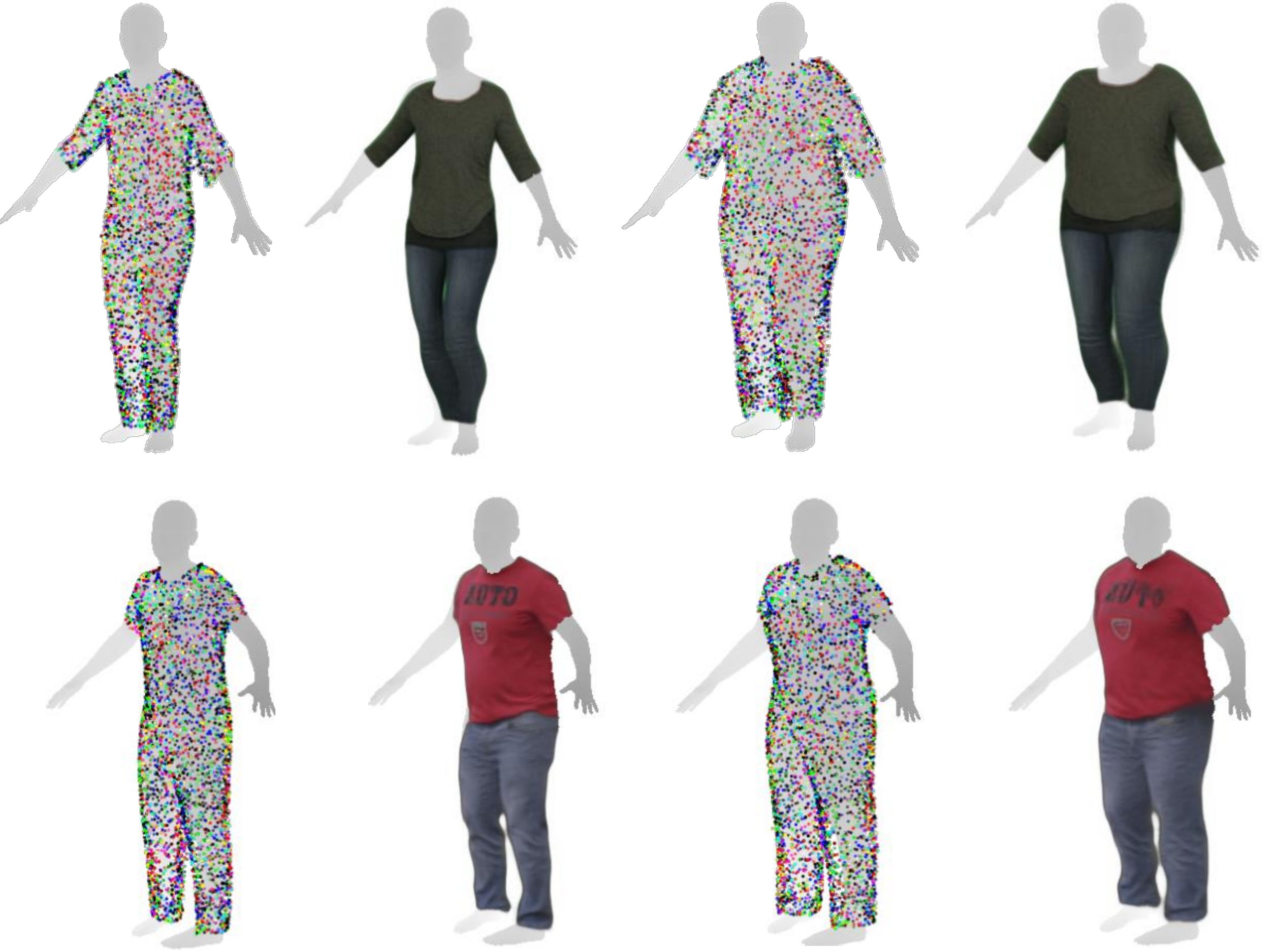}
    \caption{Our approach can also retarget the geometry and the appearance to new body shapes. The appearance retargeting works well for uniformly colored clothes, though detailed prints (e.g.~chest region in the bottom row) can get distorted.}
    \label{fig:weight}
    \vspace{-24pt}
\end{figure}
The results of the comparison are shown in Table~\ref{tab:rgbcomp}, while qualitative comparison is shown in \fig{rgbcomp}. In \fig{garment}, we show additional results for our methods. Specifically, we show a number of clothing outfits of varying topology and type that are retargeted to new poses from both test datasets. Finally, in \fig{weight}, we show examples of retargeting of outfit geometry and appearance to new body shapes within our approach.

\begin{table}[ht]
    \centering
    \resizebox{0.46\textwidth}{!}{
    \begin{tabular}{cccc}
                                     &  LPIPS$\downarrow$&  SSIM$\uparrow$ & MS-SSIM$\uparrow$ \\ \hline
                                     & \multicolumn{3}{c}{\textit{PeopleSnapshot}} \\
         Ours & \textbf{0.031} & \textbf{0.950} &  \textbf{0.976} \\ 
         StylePeople & 0.0569 & 0.938 & 0.972 \\ \hline
                                    & \multicolumn{3}{c}{\textit{AzurePeople}} \\
         Ours & \textbf{0.066} & \textbf{0.925} & 0.937 \\ 
         StylePeople & 0.0693  & 0.923 & \textbf{0.946}  \\ \hline
    \end{tabular}
    }
    \caption{Quantitative comparisons with the StylePeople system on the two test datasets using common image metrics. Our approach outperforms StylePeople in most metrics thanks to more accurate geometry modeling within our approach. This advantage is validated by visual inspection of quantitative results (\fig{rgbcomp}). }
    \label{tab:rgbcomp}
    \vspace{-10pt}
\end{table}

\section{Summary and Limitations}
We have proposed a new approach to human clothing modeling based on point clouds. We have thus built a generative model for outfits of various shape and topology that allows us to capture the geometry of previously unseen outfits and to retarget it to new poses and body shapes. The topology-free property of our geometric representation (point clouds) is particularly suitable for modeling clothing due to wide variability of shapes and composition of outfits in real life. In addition to geometric modeling, we use the ideas of neural point-based graphics to capture clothing appearance, and to re-render full outfit models (geometry + appearance) in new poses on new bodies.
% \vspace{-4pt}
\paragraph{Geometry limitations} Our model does not consider cloth dynamics, and to extend our model in that direction some integration of our approach with physics-based modeling (e.g.~finite elements) could be useful. Also, our model is limited to outfits similar to those represented in the Cloth3D dataset. Garments not present in the dataset (e.g.~hats) can not be captured by our method. This issue could be possibly addressed by using another synthetic datasets on par with Cloth3D, as well as using real-word 3D scan datasets with ground truth clothing meshes.
% \vspace{-15pt}
\paragraph{Appearance limitations} Our current approach to appearance modeling requires a video sequence in order to capture outfit appearance, which can be potentially addressed by expanding the generative modeling to the neural descriptors in a way similar to generative neural texture model from~\cite{Iskakov21}. We also found the results of our system to be prone to flickering artifacts, which is a common issue for neural rendering schemes based on point clouds~\cite{npbg}. We believe, those artifacts may be alleviated by introducing more sophisticated rendering scheme or by using denser point clouds.

%\section*{Acknowledgements}

%\FloatBarrier
%\ifnum\value{page}>8 \errmessage{Number of pages exceeded!!!!}\fi

{\small
\bibliographystyle{ieee}
\bibliography{refs}
}
\end{document}